\documentclass[10pt,twocolumn,letterpaper]{article}
\pdfoutput=1

\usepackage{cvpr}
\usepackage{times}
\usepackage{epsfig}
\usepackage{graphicx}
\usepackage{amsmath}
\usepackage{amssymb}

\usepackage{graphicx}
\usepackage{gensymb}
\usepackage{nopageno} % get rid of page numbers
\usepackage{stfloats}
\usepackage{lipsum}% http://ctan.org/pkg/lipsum
\usepackage{ textcomp } % for ~ symbol

\cvprfinalcopy % *** Uncomment this line for the final submission

\def\cvprPaperID{5928} % *** Enter the CVPR Paper ID here

\ifcvprfinal\fi
\begin{document}

%%%%%%%%% TITLE
\title{Deeper Interpretability of Deep Networks}
%Deeper Interpretability of Deep Networks, from Hidden Layers to Categorization Response
\author{
Tian Xu$^{1}$ \qquad Jiayu Zhan$^{1}$ \qquad Oliver G.B. Garrod$^{1}$ \qquad Philip H.S. Torr$^{2}$\\ 
Song-Chun Zhu$^{3}$ \qquad Robin A.A. Ince$^{1}$ \qquad Philippe G. Schyns$^{1}$\\
$^{1}$ University of Glasgow, Glasgow, UK\\
$^{2}$ University of Oxford, Oxford, UK\\
$^{3}$ University of California, Los Angeles, USA
}

\maketitle
%\thispagestyle{empty}

%%%%%%%%% ABSTRACT
\begin{abstract}
Deep Convolutional Neural Networks (CNNs) have been one of the most influential recent developments in computer vision, particularly for categorization \cite{lecun2015deep}. There is an increasing demand for explainable AI as these systems are deployed in the real world. However, understanding the information represented and processed in CNNs remains in most cases challenging. Within this paper, we explore the use of new information theoretic techniques developed in the field of neuroscience to enable novel understanding of how a CNN represents information. We trained a 10-layer ResNet architecture \cite{he2016deep} to identify 2,000 face identities from 26M images generated using a rigorously controlled 3D face rendering model that produced variations of intrinsic (i.e. face morphology, gender, age, expression and ethnicity) and extrinsic factors (i.e. 3D pose, illumination, scale and 2D translation).  With our methodology, we demonstrate that, unlike humans, the network overgeneralizes identities even with extreme changes of their face shape, but it is more sensitive to changes of texture.  To understand the processing of information underlying these counterintuitive properties, we visualize the features of shape and texture that underlie identity decisions.  Then, we shed a light of information processing into the black box and demonstrate how the hidden layers represent features for decision, and characterize the invariance of these representations to changes of 3D pose.  We hope that our methodology will provide an additional valuable tool for interpretability of CNNs.
\end{abstract}

%%%%%%%%% BODY TEXT
\section{Introduction}

Hierarchical CNNs and their multiple nonlinear projections of visual inputs have become prime intuition pumps to model visual categorization in relation to the hierarchical occipito-ventral pathway in the brain (\cite{cadieu2014deep, cichy2016comparison, khaligh2014deep, VanRullen2017, Schrimpf407007}).  However, understanding the information represented and processed in CNNs is a cornerstone of the research agenda whose resolution would enable more effective network designs (e.g. by using CNNs as modular building blocks that can perform specific functions), more robust practical applications (e.g. by predicting adversarial attacks) and broader usage (e.g. as information processing models of the brain).  Here, we developed a new methodology to address the deeper interpretability of the information processing mechanisms of CNNs and testing its applicability in a case study. 

A starting point to understand information processing in CNNs (and the brain) is to identify the features represented across their respective computational hierarchies. In CNNs, multi-layered deconvolution techniques (deconvnet) \cite{Zeiler2014} can identify features of increasing complexity and receptive field size represented in the lower convolution layers to the mid and higher-level layers. In the brain, reverse correlation has been successfully applied to visualize the receptive fields of different brain regions along the occipito-ventral hierarchy \cite{popivanov2016stimulus, schyns2007dynamics, ringach2004mapping, hubel1962receptive, zhan2018dynamic}. An open question remains whether a well-constrained CNN (i.e. constrained by architecture, time, representation, function and so forth) could learn the mid-to-high-level features that flexibly represent task-dependent visual categories in the human visual hierarchy.

To evaluate the usefulness of CNNs as models of brain computations, researchers can quantify their predictive power for neural responses (i.e. how accurately hidden layers predict the activity of specific brain regions), and also assess the algorithmic understanding they enable (i.e. the information processing light they shed on computations in brain networks).  To quantify predictive power, researchers can compute the similarity between the activity of CNNs' hidden layers and that of brain regions in response to the same stimulus categories (e.g. \cite{cadieu2014deep, cichy2016comparison}. However, a deeper similarity of computation is necessary to use CNNs as understandable models of the underlying visual categorization mechanisms. Without such deeper understanding of information processing, all that CNNs offer are layered silicon black boxes to predict the performance of the layered wet ones, not to explain how brains achieve these categorizations across the occipito-ventral hierarchy.

\begin{figure*}[htbp]
\center
\includegraphics [width=\textwidth] {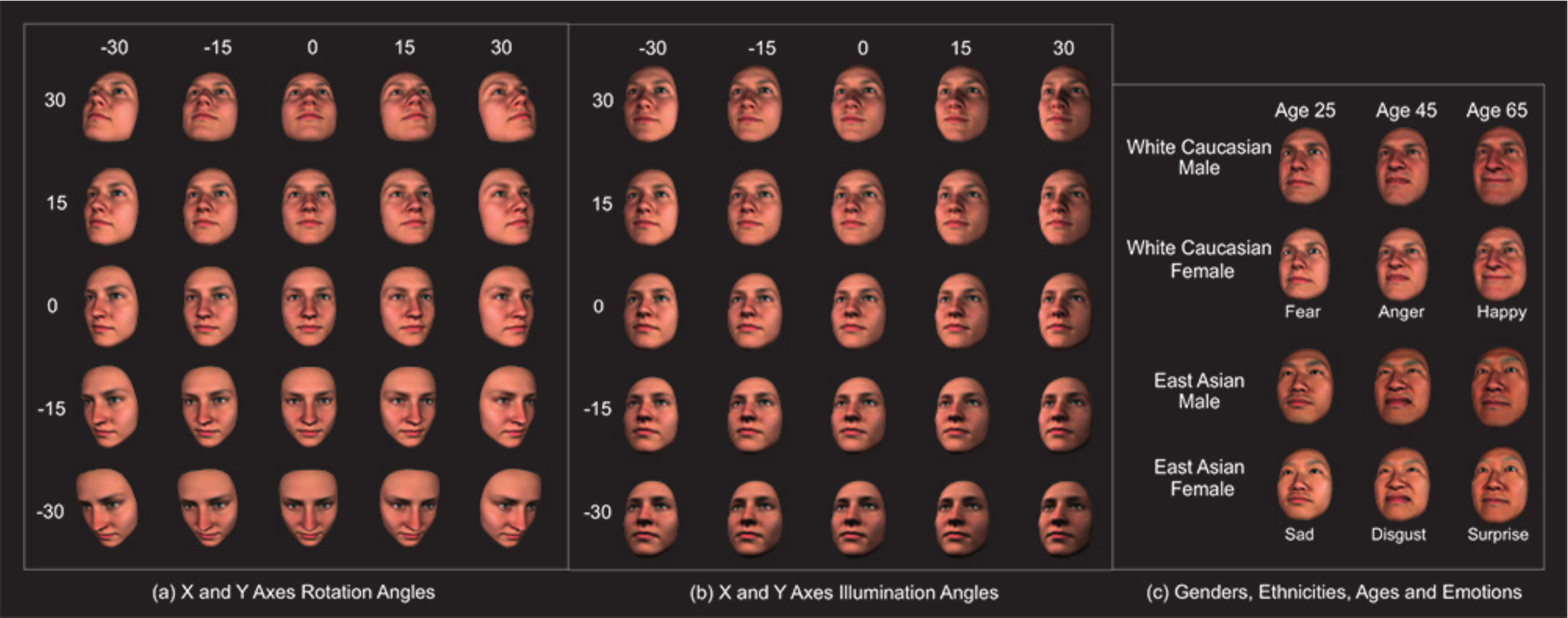}  %[scale=1.5]. \textwidth
\caption{Extrinsic (a and b) and Intrinsic (c) Factors of Variance of the Generative Model of 3D Faces. We generated 2,000 different identities and used the extrinsic and intrinsic factors of face variance to combinatorially generate 26M face images.}
\label{fig1}
\end{figure*}

In both CNNs and the brain, we need to model how stimulus information is transformed across hidden layers and brain regions to produce task-dependent responses in the hierarchy that ultimately lead to categorization response. Our main contribution to this huge challenge is to propose a new psychophysical methodology based on information theory with which we could understand how the brain reduces the high dimensional visual input to the low dimensional features that support distinct behaviors \cite{zhan2018dynamic}.  Its key feature is to better control of stimulus variation to understand the stimulus information underlying CNN categorization responses, and its transformations across the layers. Thus, rather than using an existing database of varied images from multiple natural categories and benchmark CNN performance, we rigorously controlled the factors of image generation using a single stimulus category and task---i.e. faces and their identification.  Our approach enables a deeper understanding of the information processing within CNNs, which in turn enables their usage as understandable information processing models of the brain \cite{Kay2017, zhan2018dynamic}.

\section{Related Work}
\begin{figure*}[t]
\center
\includegraphics [width=0.95\textwidth] {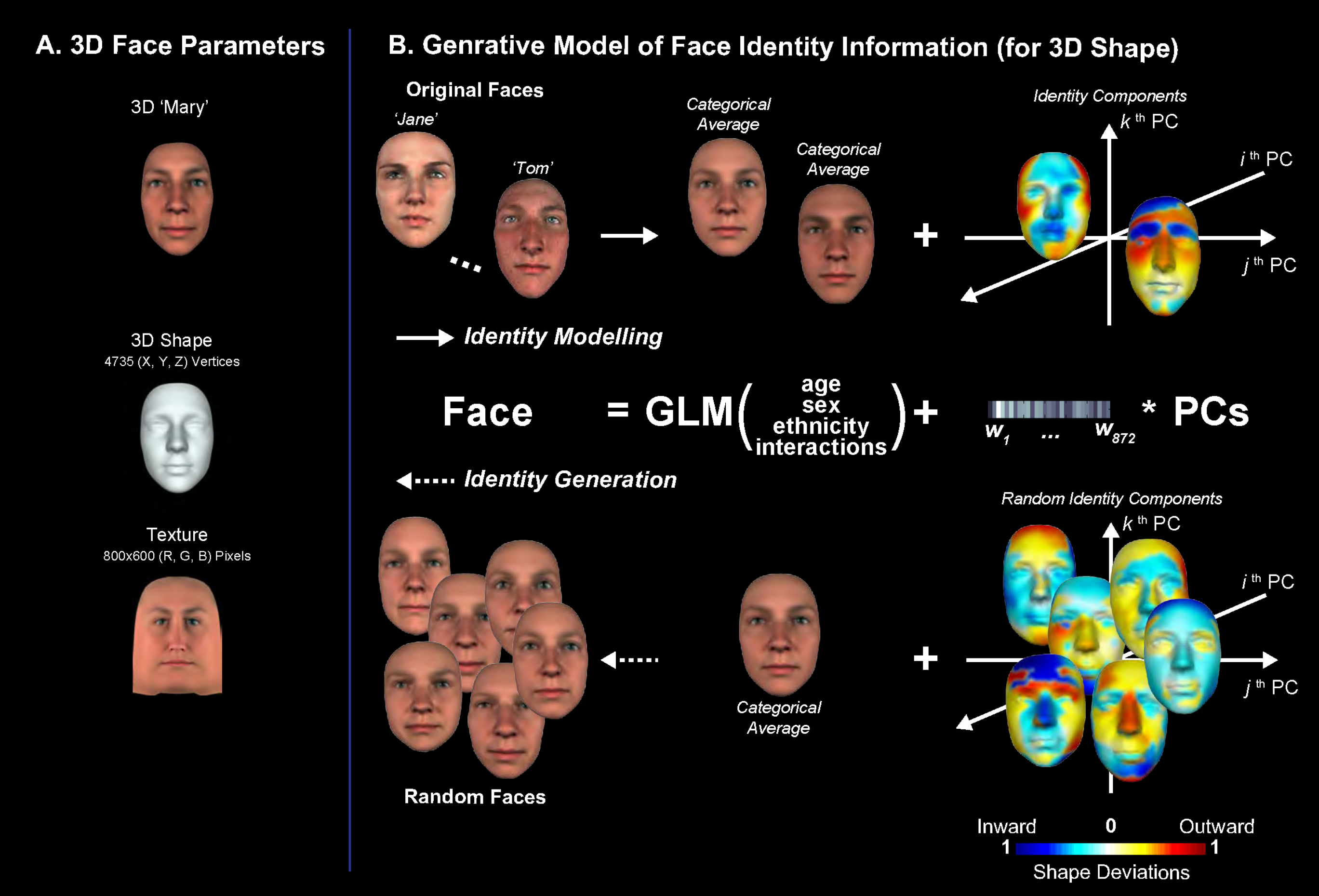}  %[scale=1.5]. \textwidth
\caption{Generalized Linear Model of Face Identity - Random Identity Generation.  A. A given 3D face identity comprised random multivariate shape and texture information dimensions. B.  We constructed a generative model by applying a Generalized Linear Model, independently for shape and texture, to a database of 3D scanned faces.  We extracted the variance associated with the intrinsic factors face age, sex, ethnicity and their interactions, leaving out identity residuals for each scanned face (illustrated only for shape).  We applied Principal Components Analysis to the residuals.  In generative mode, to produce one random face identity, we multiplied a random vector defining each random identity by the principal components of identity residuals, to create random identity residuals which were then added to the categorical average.}
\label{fig2_1}
\end{figure*}

Face categorization is an important benchmark in human and machine vision research, because is a well constrained stimulus class that nevertheless conveys a wealth of different social signals that can be mathematically modelled for real world applications \cite{o2018face, jack2017toward}. 

In human vision, the challenge is to understand where, when and how information processing mechanisms in the brain realize face identification, given extensive image variations such as those presented in Figure \ref{fig1} (plus translation and scaling). Psychophysical reverse correlation techniques (e.g. Bubbles, \cite{gosselin2001bubbles}) enabled reconstruction of the stimulus information underlying various face recognition tasks, using either behavioral or brain measures \cite{smith2009inverse, ince2015tracing, zhan2018dynamic}. In particular, new methods can represent the features relevant to different categorization tasks and isolate the brain regions where, and when these features are combined to achieve behavior \cite{zhan2018dynamic}. Representational Similarity Analysis (RSA, \cite{kriegeskorte2008representational}) is another popular method that compares the responses of different architectures (e.g. human behavior, computational models and brain activity) to the same input stimulus categories.  In its current applications, it does not isolate the stimulus features responsible for the responses and thus does not reveal the deeper similarities of information processing that cause the responses.

In computer vision, the challenge has been to increase categorization performance using deep learning methods. The approach is to use large datasets of images (e.g. DeepFace \cite{taigman2014deepface}, FaceNet \cite{schroff2015facenet}, face++ \cite{zhou2015naive}, Labeled Faces in the Wild database (LFW) \cite{huang2007labeled}, Youtube Faces DB \cite{wolf2011face}) and demonstrate that well designed and trained deep neural network can outperform humans \cite{o2018face}.  However, understanding the information processing underlying their high performance levels remains a challenge that must be resolved to address the shortcomings revealed by adversarial testing. There is therefore a strong focus on better understanding CNNs.  For example, Zeiler and Fergus  \cite{Zeiler2014} famously used deconvolutional networks to identify the image patches responsible for patterns of activation.  Relatedly, Simonyan et al. \cite{simonyan2013deep}'s visualization technique based on gradient ascent can generate a synthetic image that maximally activates a deep network unit.  The Class Activation Maps (CAM) of Zhou et al. \cite{zhou2016learning} can highlight the regions of the image the network uses to discriminate \cite{o2018face}.  \cite{ribeiro2016should} built a locally interpretable model around a particular stimulus, to determine the parts of the image (or words of a document) that are driving the model's classification.

Here, to develop an understandable AI of CNNs and understand their inner information processing, we examined the relationships between three classes of variables: stimulus feature dimensions, hidden layer responses and output responses \cite{zhan2018dynamic}.  This is a different approach to typical CNN research because we aim to: (1) isolate and control the main factors of stimulus variance to (2) precisely measure the layer-by-layer co-variations of these factors that influence network output responses. Such tight psychophysical control is difficult to achieve with the large datasets of unconstrained 2D images.

\section{Generative Model for 3D Faces}
To achieve these goals, we used a generative model of the face information that controls and tests the effect of each factor of face variance (i.e. the objectively available information) on CNNs' performance. Though Generative Adversarial Networks (GANs) \cite{goodfellow2014generative} provide image-to-image translation (e.g. CycleGAN \cite{zhu2017unpaired} and StarGAN \cite{choi2017stargan}) with reasonable quality, and so can be treated as generative models, they do not explicitly characterize the generic generative parameters of the translated image (e.g. parameters for 3D face shape). 

Our Generative Model of 3D Faces (GMF) \cite{yu2012perception, zhan2017efficient} mixed explicitly defined and latent generative parameters to generate 2,000 face identities with intrinsic variance factors of 500 random face variants $\times$ 2 genders $\times$ 2 ethnicities $\times$ 3 age (25, 45 and 65 years) and $\times$ 7 emotions (i.e. ``happy'', ``surprise'', ``fear'', ``disgust'', ``anger'', ``sad'' and ``neutral''). Each of these combinations was further varied according to extrinsic factors of rotation and illumination (both ranges from -30$\degree$ to +30$\degree$ by increments of 15$\degree$) along the X and Y axes to produce a controlled database of a total of 26M images. Figure \ref{fig1} illustrates the extrinsic and intrinsic variations for one example of face morphology.

Figure \ref{fig2_1} illustrates the image generation using 3D face shape (face texture is separately and similarly handled).  Briefly, (see Zhan et al. \cite{zhan2017efficient} for details), a Generalized Linear Model (GLM) extracted the explicit factors of age, ethnicity, gender and their interactions from a database of 872 scanned real 3D faces while the remaining unexplained part of each identity was controlled by the principal components of the GLM residual matrix. To generate 2,000 random identities, we inverted the model, and multiplied the principal components of identity residuals with 500 random coefficients vectors to produce a distinct residual identity vector that defines each face identity which was then added to all permutations of the GLM factors (for a total of 2,000 identity vectors computed separately for shape, as shown in Figure \ref{fig2_1}, and texture, not shown). In Section 5 below, we added noise to the shape and texture vectors defining each identity and generated face images to test network performance.

\section{CNN: 10-layer ResNet}

\begin{figure}[b] %[htbp]
\center
\includegraphics [scale=0.6] {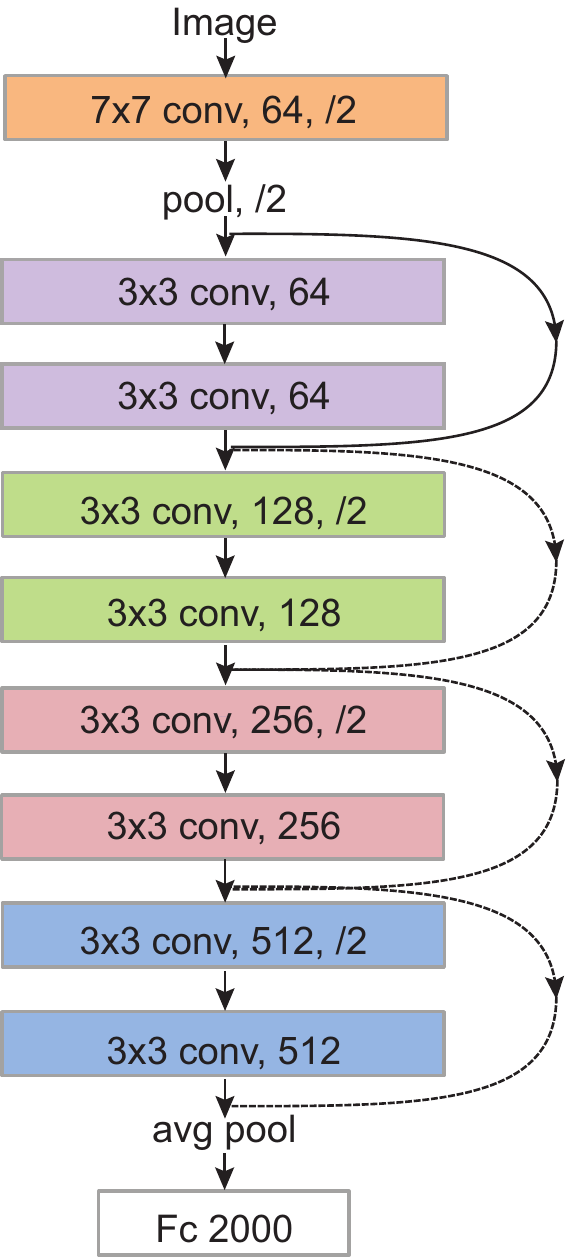}  %[scale=1.5]. \textwidth \columnwidth
\caption{10-layer ResNet Architecture}
\label{fig2}
\end{figure}

A 10-layer ResNet learned to associate the face images with their identity. We chose ResNet because it is a state-of-the-art architecture that achieved high classification performance on various datasets. We used only 10 layers (ResNet-10) to keep network complexity relatively low for the analysis of hidden layers detailed later.

We applied the training and testing regime of \cite{xu2018using}. For training, we randomly selected 60\% of the generated face images, for a total of 15,750,000 images. At training, we applied data augmentation to increase data complexity and to alleviate overfitting by randomly scaling (between 1$\times$ and 2$\times$) and translating images in the 2D plane (between 0 and 0.3 of the total image width and height). At testing, we used the remaining 40\% images, for a total of 10,500,000 images.

\section{Generalization and Adversarial Testing}
\begin{figure*}[t] %[htbp]
\center
\includegraphics [width=\textwidth] {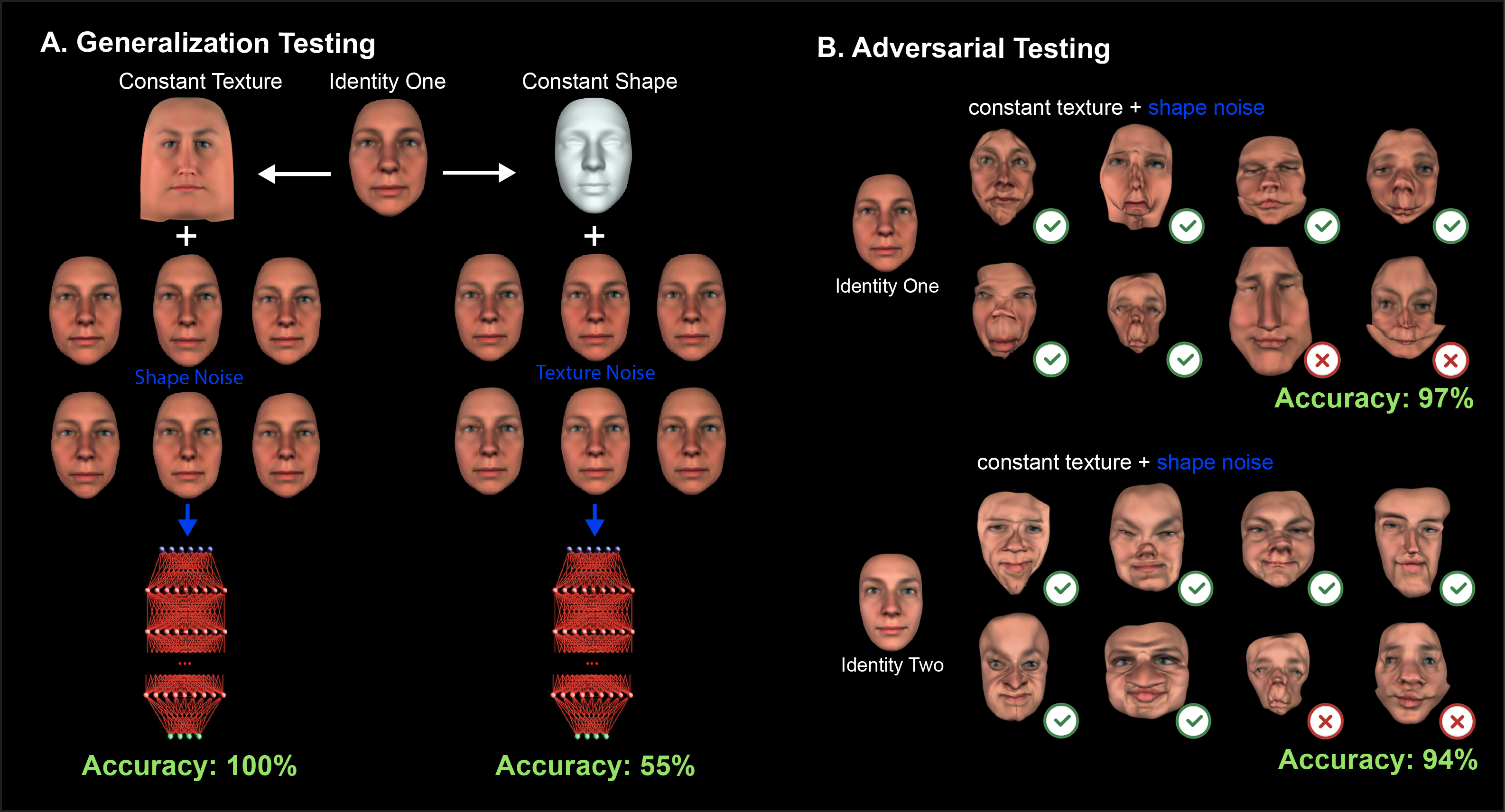}  %[scale=1.5]. \textwidth
\caption{A.  We added multivariate noise (shape, left, or texture, right panel) while keeping texture (left) vs. shape (right panel) constant and measured network identification accuracy in response to both.  B.  Adversarial testing with addition of extreme shape noise nevertheless revealed high identification accuracy, but with generalization to grotesque faces for both identities.}
\label{fig4}
\end{figure*}

ResNet correctly identified the testing image set with very high accuracy (i.e. over 99.9\% across all variations of intrinsic and extrinsic factors, which would easily outperform humans using the same dataset \cite{o2018face}), though it was sensitive to the similarity of training data \cite{xu2018using}.

\subsection{Generalization Testing}

A powerful methodology to test the boundaries of human categorization performance is to add or multiply the original stimulus with noise \cite{murray2011classification, jack2017toward}.  We applied a similar approach with ResNet, by adding noise directly into the generative model. Specifically, we added a random vector of multivariate Gaussian noise with diagonal covariance (separately for 3D shape and 2D texture) to the vector defining the identity of a face in the generative model.  This produced stimulus variations in shape (and texture) around this identity.  We kept the noise level at a 0.8 proportion of the coefficients defining the identity.  In Figure \ref{fig4}A, the left (vs. right) column of images illustrates the top face identity with added shape (vs. texture) noise, while keeping its texture (vs. shape) constant. Note that whereas variations in shape (left panel) look like slightly different face identities to human observers, variations in texture (right panel) do not apparently change the identity of the face \cite{zhan2017efficient, nemrodov2019multimodal}. We generated 10,000 such noisy variations of shape and texture for two example face identities and rendered them as testing images.

ResNet identified 3D shape variations with 100\% performance, whereas varying texture incurred a performance drop \texttildelow55\%. Unlike humans \cite{zhan2017efficient, nemrodov2019multimodal}, ResNet's high performance relies on 2D face texture more than on 3D shape.

\subsection{Adversarial Testing}

To illustrate this counterintuitive performance, we adversarially tested ResNet with a 3D shape noise level 5 times higher than that defining each random identity, while leaving texture unchanged. Using 1,000 such adversarial faces for each identity, ResNet nevertheless over-generalized them as the target identity (at 97\% and 94\% performance, respectively).  Figure \ref{fig4}B reveals several examples of over-generalized grotesque faces (see green tick signs; red cross signs represent a rejection of the distorted face as an exemplar of the target identity). Adversarial testing compellingly illustrates that 3D face shape information is less important to ResNet than texture. It also demonstrates that our network would fail face identification tasks where immunity to adversarial exemplars is critical.

\section{Information Representation for ResNet Decision}

\begin{figure*}[htbp] %[htbp]
\center
\includegraphics [width=0.75\textwidth] {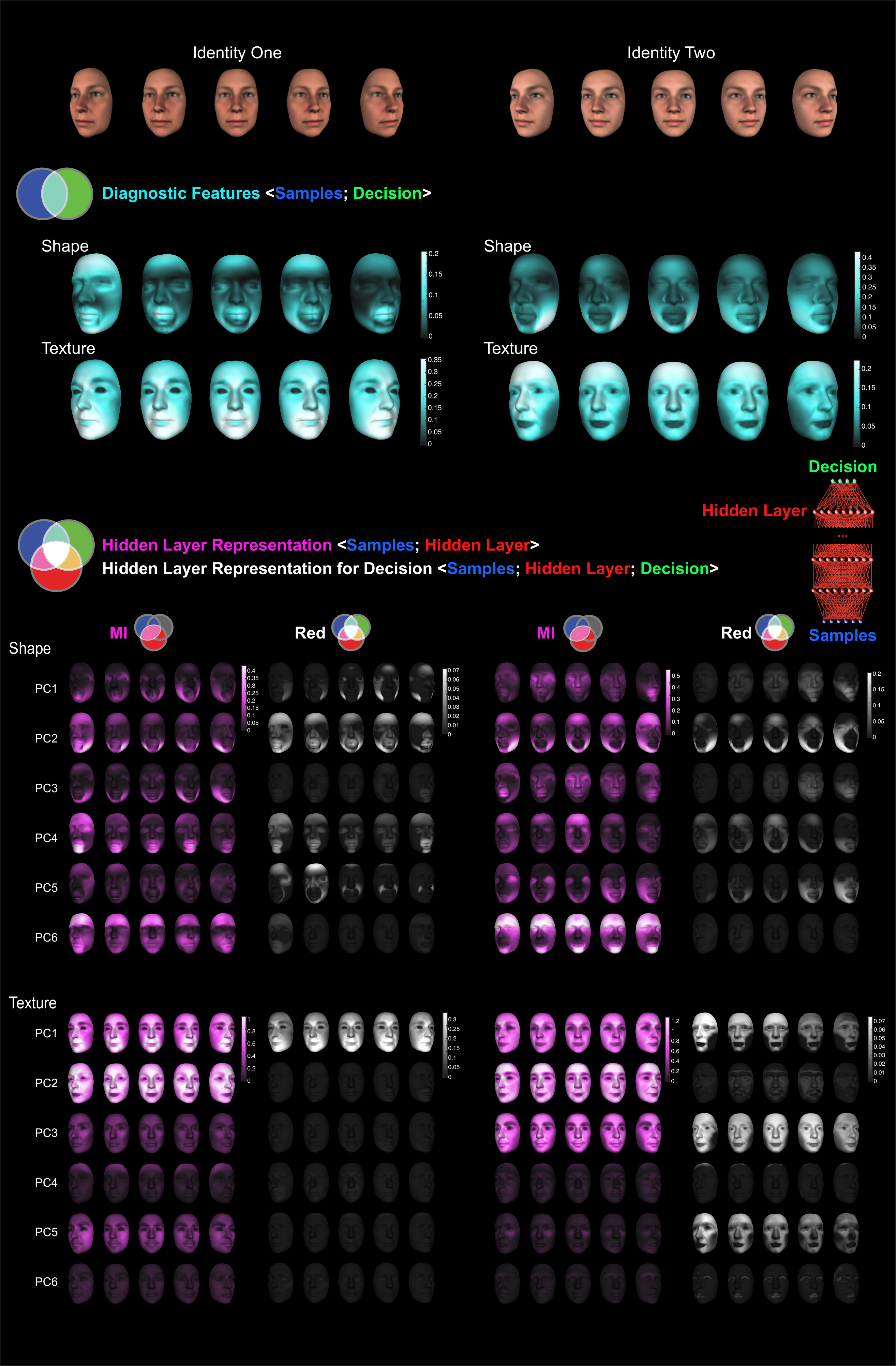}  %[scale=1.5]. \textwidth
\caption{\underline{Diagnostic Features (cyan intersection and face features).}  For each 3D shape vertex and 2D RGB pixel and each face orientation of two face identities, we computed the Mutual Information between their variations due to multivariate noise and the real-valued identity response of ResNet on its top layer.  This analysis informed the diagnostic information that the network must process in its hidden layers between stimulus and response.  \underline{Hidden Layer Representation (magenta intersection and face features).}  For shape and texture information, mutual information reveals, for the first 6 principal component (PC) of Resnet layer 9.5 activity, the face features represented.  \underline{Hidden Layer Representation for Decision (white intersection and face features).}  For shape and texture information, mutual information reveals, for the first 6 principal component (PC) of Resnet layer 9.5 activity, the subset of face features represented for decision.}
\label{fig5}
\end{figure*}

Using again the multivariate noise procedure, we derived a deeper interpretability of the layers of ResNet, starting with its top decision layer---i.e. its categorization behavior.  Across testing trials, noise introduce variations in the 3D location of each shape vertex and in the RGB values of each 2D texture pixel. ResNet responds to these variations both in its hidden layers, and on its output layer.  We first explain how we visualized the shape and texture features that modulated output unit responses.  Following this, we apply a similar analysis to the hidden layers.

The input variations due to noise produced real-valued variations of the output unit that responds maximally to the targeted identity---i.e. before $Argmax$ on calculated across the 2,000 units of the decision layer.  To visualize the shape vertices and texture pixels that modulate ResNet output response, we computed with Mutual Information (MI) the relationship between stimulus variation ($S$) and output unit response ($R$) using a semi-parametric lower bound estimator (Gaussian-Copula Mutual Information, GCMI, \cite{ince2017statistical}):
\begin{equation}
MI(S;R) = H(S) + H(R) - H(S;R)
\end{equation}
GCMI identifies vertices and pixels that affect response for a given identity (i.e. the diagnostic vertices and pixels).  GCMI therefore reveals the stimulus features the network must necessarily process, between the input faces and their identification on the decision layer.  In our methodology, we illustrate diagnostic information as the set cyan intersection between input information samples (the blue set) and the corresponding output decision responses (the green set, see Figure \ref{fig5}).

Figure \ref{fig5} shows the cyan diagnostic information reported on the two example identities.  That is, the shape and texture features that support the network decisions for identity 1 (e.g. 3D vertices around the jaw line, mouth and forehead; 2D RGB pixels around the mouth) and for identity 2 (i.e. 3D vertices forming the cheeks and the forehead texture). 

We repeated this analysis across the 5 face viewpoints the network was trained on and found usage of the same diagnostic face features across viewpoints---i.e. viewpoint-invariance of the diagnostic features.  Know what features ResNet uses to identify the two faces, we now track the organization of feature representation (diagnostic and not) in the hidden layers.

\section{Information Representation in the Hidden Layers}
\begin{figure*}[htbp] %[htbp]
\center
\includegraphics [width=\textwidth] {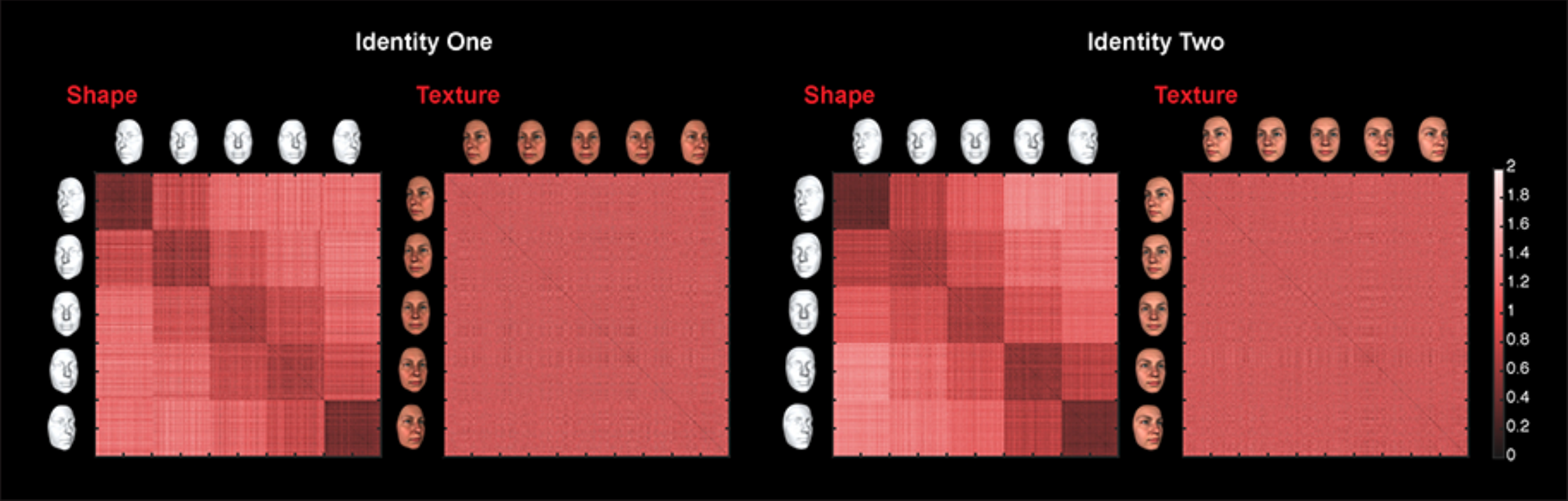}  %[scale=1.5]. \textwidth
\caption{Dissimilarity analysis of hidden layer 9.5 in ResNet.  We organized the PC scores response of the layer to the face identity inputs by their orientation in depth (5 orientations, from -30$\degree$ to + 30$\degree$ with 15 $\degree$ increments, with 10,000 noisy face exemplars per orientation and identity).  The two identities demonstrate viewpoint-dependent responses of layer 9.5 to shape, and viewpoint-invariant response to texture. }
\label{fig6}
\end{figure*}

We analyzed hidden layer representations, starting one layer down from the response layer (i.e. the average pooling layer after layer 9, henceforth called ``layer 9.5'').  First, we computed the multivariate activation of layer 9.5 by feeding ResNet with the 10,000 shape and 10,000 texture variations images for each identity and viewpoint used earlier.  We reduced the dimensionality of the multivariate activation with a randomized Principal Components Analysis (PCA) algorithm \cite{halko2011finding, li2017algorithm} computed separately for each combination of identity and their shape and texture variations. This stage produced 4 matrices of 50,000 PCs score vectors (5 viewpoints x 10,000 variations) for each combination of identity and their shape and texture variations. 

\subsection{Property of Representations in the Hidden Layers}

Remember that the output layer of ResNet responds to the same shape and texture face features across viewpoints (cf. Section 6).  Here, we asked whether the activation of layer 9.5 represents viewpoint.  To this end, we ordered the 4 matrices of PC scores vector by the 5 face viewpoint (i.e.  -30$\degree$ to + 30$\degree$, by 15 deg increments).  We computed a dissimilarity matrix \cite{kriegeskorte2008representational} by cross-correlating the 50,000 PC score vectors---using the dissimilarity measure (1 - Pearson correlation) between any pair of score vectors.  Figure \ref{fig6} presents the results.  For each identity, face shape elicited viewpoint-dependent activations on layer 9.5.  The dissimilarity matrices reflect such viewpoint representations with a blocked structure across the diagonal, which demonstrates that the blocks of 10,000 face images at the same orientation are represented more similarly on the hidden layer than face images at any other orientation.  In contrast, face texture elicited viewpoint invariant activations on layer 9.5.  Thus, the activity of layer 9.5 represents face orientation, but only for shape.  For texture, the network has reduced this varying input dimension in the layers underneath 9.5.

\subsection{Representations Viewpoint Dependent and Viewpoint Invariant Features in the Hidden Layers}

We now know from the dissimilarity analysis of layer 9.5 activations that it differently represents face shape and texture.  However, we still do not know which specific face shape and face texture features represented on layer 9.5 underlie the reported viewpoint-dependent/invariant performance. To derive such a deeper understanding of the information processed, we focussed our analysis on the first 6 principal components of activation of layer 9.5 that explain (26\% - 30\%) of this layer's activation variance.

We used again GCMI to quantify the relationship between the 10,000 input variations (in shape and texture) and the corresponding variable activations of the 6 PCs, separately at each of the five orientations (i.e. 10,000 trials per orientation).  This analysis reveals all the shape and texture face features represented on layer 9.5.  However, a subset of these features (the diagnostic features) are used by layer 10 for the final classification output. To dissect the diagnostic from the nondiagnostic features represented on layer 9.5, we repeated the analysis, substituting GCMI with information theoretic redundancy.  Redundancy quantifies how the samples ($S$) (i.e. variations of each 3D shape vertex and RGB texture pixels) are co-represented (i.e. redundantly represented) in layer 9.5 activity ($L$) and output response ($R$).  Formally, redundancy ($Red$) is the intersection of two mutual information quantities as shown below  \cite{mcgill1954multivariate, bell2003co, ince2017measuring}:
\begin{equation}
Red(S;L;R) = MI(S;L) + MI(S;R) - MI(S; R,L)
\end{equation}
We compared the feature representations derived with GCMI and redundancy on layer 9.5 to understand how this layer selects and inherits shape and texture for final decision on output layer 10. In Figure \ref{fig6}, the magenta GCMI faces demonstrate that the layer represents many different shape and texture features on its PCs. In contrast, the white faces computed with redundancy from the same PCs directly visualize the subset of shape and texture features represented for decision.  In Figure \ref{fig5}, we can now compare the three critical classes of features derived in our framework (they represented as three colored set intersections): Namely, the cyan features of the top decision layer, the magenta GCMI features of layer 9.5 and the redundant white features of features.  They reveal that only a subset of the shape and texture features represented in layer 9.5 (see magenta faces) are used by ResNet for final decision (see white and cyan faces, respectivey): for Identity One: primarily the white PC2 (shape) and PC1 (texture): for Identity Two primarily white PC2 (shape) and PC3 (texture). 

In sum, Figure \ref{fig5} demonstrates how mutual information and redundancy methods can assist the interpretation of the hidden layers of deep networks, by separating information represented on a given layer that affects categorization response from that which does not (see \cite{zhan2018dynamic} for a similar dissociation in brain representations). The methodology can be also extended to other layers to understand the information flow within the network.

\section{Conclusion and Discussion}
We trained a deep network on a controlled set of face images and found that it behaved dramatically differently to human face perception: performance was almost invariant to shape deformations, while being extremely sensitive to variations of texture.  We achieved a deeper interpretation of the network with a methodology that tightly controls the generative dimensions of the tested visual category. Following learning of varying but controlled face identity images, we used psychophysical testing with targeted multivariate noise (i.e. noise on the generative dimensions defining the face identity).  

We applied information theoretic measures to the triple $\langle$samples; hidden layer Response; Decision$\rangle$ and made several important new findings.  First, we visualized the specific diagnostic shape and texture features the network uses to identify faces.  Second, using redundancy we tracked the representation of diagnostic features in a hidden layer, separating it from other represented features.  Finally, we dissociated properties of viewpoint-dependent representation of shape features from viewpoint-invariant representation of texture features, on the same hidden layer. We believe such deeper understanding of information processing in deep networks is now necessary to start establishing their algorithmic similarities to other architectures (e.g. brains or other networks).  Our methodology can be extended to measure the relationship between input information samples and its representation in the layers of architecture 1 and architecture 2 as $Red$ $\langle$Samples; layer architecture 1; layer architecture 2$\rangle$.  It could also be fruitfully applied to better understand the information causes of adversarial attacks and, with further developments, to build CNN modules that perform specific functions on their inputs (e.g. a face identifier, pose identifier and so forth).

Building from our work, the main challenge to further a deeper information processing understanding of CNNs is to better control the information they learn so we can test how it is represented and transformed in the network for various output responses.  This can be achieved with two main approaches:  First, by directly engineering new generative models of face, object and scene categories that faithfully reflect the statistics of real-world faces, object and scenes \cite{qi2018human}.  Second, by indirectly modelling (e.g. with CNNs) the latent generative factors of very large databases of face, object and scene images \cite{ghosh2017multi}.  As with understanding information processing in the brain, we will only get out of CNNs what we put in.

%\section*{Acknowledgements}
%PGS is funded by the Wellcome Trust (107802/Z/15/Z) and the Multidisciplinary University Research Initiative (MURI) / Engineering and Physical Sciences Research Council (EP/N019261/1).

%%%%%%%

%{\small
%\bibliographystyle{ieee}
%\bibliography{psy}
%}

\end{document}